\documentclass{article}
\usepackage[preprint]{log_2026}			

\usepackage{amsmath}						
\usepackage{amssymb}
\usepackage{booktabs}						
\usepackage{enumitem}
\usepackage{graphicx}						
\usepackage{makecell}                       
\usepackage{tabularx}						

\usepackage[
	backend=biber,
	style=numeric-comp,
	backref=true,
	natbib=true]{biblatex}
\DeclareSourcemap{\maps[datatype=bibtex]{
	\map{
		\step[fieldsource=doi, final]
		\step[fieldset=issn, null]
		\step[fieldset=url, null]
	}
}}

\definecolor{FirstPlace}{rgb}{0.2, 0.6, 0.2}
\definecolor{SecondPlace}{rgb}{0.8, 0.8, 0.2}

\newcommand{\first}[1]{\textcolor{FirstPlace}{\textbf{#1}}}
\newcommand{\second}[1]{\textcolor{SecondPlace}{\textbf{#1}}}

\newcolumntype{C}[1]{>{\centering\arraybackslash}p{#1}}

\title[\texttt{DCBA} -- Graph to Config]{Graph Data Augmentation via Contrastive Generator Inversion (\texttt{DCBA})}

\author[Stolarski et al.]{%
Mateusz Stolarski\thanks{Equal contribution.}\\
{\scriptsize Wroc{\l}aw University of Science and Technology}\\
{\scriptsize University of Technology Sydney}\\
\email{mateusz.stolarski@pwr.edu.pl}
\And
Micha{\l} Czuba\footnotemark[1]\\
{\scriptsize Wroc{\l}aw University of Science and Technology}\\
\email{michal.czuba@pwr.edu.pl}
\And
\L{}ukasz Krai\'{n}ski\footnotemark[1]\\
{\scriptsize SGH Warsaw School of Economics}\\
\email{lkrain@sgh.waw.pl}
\And
Katarzyna Musial\\
{\scriptsize University of Technology Sydney}\\
\email{katarzyna.musial-gabrys@uts.edu.au}
\And
Pawe\l{} Pra\l{}at\\
{\scriptsize Toronto Metropolitan University}\\
\email{pralat@torontomu.ca}
\And
Bogumi\l{} Kami\'{n}ski\\
{\scriptsize SGH Warsaw School of Economics}\\
\email{bogumil.kaminski@sgh.waw.pl}
\And
Piotr Br{\'o}dka\\
{\scriptsize Wroc{\l}aw University of Science and Technology}\\
{\scriptsize University of Technology Sydney}\\
\email{piotr.brodka@pwr.edu.pl}
}

\begin{document}

\maketitle

\begin{abstract}
Graphs provide a natural representation of many complex systems, ranging from social platforms to ecosystems. However, the development of graph-based machine learning methods is often constrained by the limited availability of large and diverse graph datasets. In this paper, we introduce \texttt{DCBA}, a model-based approach to graph data augmentation that infers the configuration of a synthetic graph generator from an observed network. We instantiate the proposed framework using the \texttt{ABCD} generator, which produces scale-free networks with community structure. Our model learns a joint representation of graphs and generator parametrisations using a multi-positive contrastive objective with soft negative weighting. The learned representation enables the prediction of an \texttt{ABCD} configuration whose stochastic realisations preserve the macrostructural properties encoded by the generator. Experiments show that \texttt{DCBA} recovers generator parameters more accurately and robustly than an algorithmic inverse-modelling baseline. Its downstream utility is further demonstrated in community detection, where inferred configurations used to fine-tune \texttt{PRoCD} improve AMI on average by $161\%$ on synthetic and $273\%$ on real-world networks.
\end{abstract}



\section{Introduction}

Among the principal data domains considered within geometric deep learning~\cite{bronstein2017geometric} are graphs. They serve as a convenient means of modelling systems that depend on both individual entities and the relations between them, including social platforms~\cite{paradowski2024peer}, biological systems~\cite{jeong2001lethality}, and transport networks~\cite{cardillo2013eutransportation}. Although machine learning methods allow many problems arising in such domains to be addressed, their development remains constrained by the limited availability of large, diverse, and accessible graph datasets~\cite{zhao2023graphdataaugmentationgraph,zhou2025data}. Real-world networks may be costly to collect, restricted by privacy considerations, or available only as a single observed realisation of an underlying system.

Data augmentation is widely used to alleviate data scarcity. In computer vision, geometric transformations such as rotations or cropping can create additional samples while generally preserving their semantic content~\cite{shorten2019survey}. Such transformations are less straightforward for graph-structured data, whose semantics may depend jointly on connectivity, attributes, and higher-order structural patterns. Even small modifications can therefore remove meaningful relations or introduce implausible ones. In many applications, the desired outcome is not merely a perturbed version of the observed graph, but multiple plausible realisations that preserve selected macrostructural properties of the underlying system. This motivates performing augmentation through a fitted stochastic graph generator rather than through individual node- or edge-level operations.

\subsection{Structural Graph Augmentation}

Graph data augmentation methods can be classified according to the graph elements they modify, the operating scale, and whether the transformation is predefined or learned~\cite{zhao2023graphdataaugmentationgraph,zhou2025data}. Common structural operations include node dropping, edge perturbation, rewiring, and subgraph sampling. For example, \texttt{GraphCL} constructs contrastive views using several predefined transformations~\cite{you2020graphcontrastive}, whereas \texttt{GAug} learns task-oriented edge modifications~\cite{zhao2021dataaugmentation}. Such methods are well suited to transforming a particular observed graph, but their local modifications are task-dependent and provide limited control over macrostructural properties. They are therefore less natural when the objective is to obtain multiple realisations of the same system while preserving properties such as its degree distribution and community organisation. It is important to note that structural modifications extend beyond data augmentation; for example, in network anonymisation, the observed graph is deliberately altered to reduce node uniqueness while preserving selected properties or analytical utility~\cite{jong2026anonymization}.

\subsection{Generator-based Graph Augmentation}

Synthetic graph generators provide a natural mechanism for producing additional network realisations under explicit structural assumptions. Different families encode different properties, including homogeneous connectivity~\cite{er_model}, heterogeneous degree distributions~\cite{sf_model}, and community structure~\cite{holland1983stochastic}. Once a suitable generator and its configuration are selected, multiple stochastic realisations can be sampled while controlling the macrostructural properties represented by the model.

Previous work has already explored this idea. For instance, a framework for synthetic attributed-graph generation was introduced in~\cite{darabi2025framweork}, combining a generalised \texttt{R-MAT/SKG} structural model~\cite{chakrabarti2004r,leskovec2005realistic} with separate generators for node and edge attributes, enabling the generation of datasets substantially larger than the original input. \texttt{G-Mixup} instead estimates class-level \textit{graphons} from collections of labelled graphs, interpolates between them, and samples synthetic graphs with mixed labels~\cite{han2022gmixup}. These methods require either fitting a multi-component generative pipeline to an available dataset or access to multiple labelled graphs, and do not directly address the recovery of a reusable parametric generator configuration from a single observed network. 

\subsection{Inverse Graph Modelling}

Graph-generator calibration has previously been approached through simulation-based inference. For example, Approximate Bayesian Computation combined with spectral graph metrics can estimate generator parameters without a tractable likelihood, but requires repeated simulation and comparison for each observed network~\cite{fay2015graphmetrics}. More recently, calibration of the multilayer \texttt{mABCD}~\cite{krainski2025mabcd} generator combined direct estimators with iterative optimisation~\cite{brodka2026twinning}. This approach also highlights a limitation of estimating parameters separately: their interdependence allows errors in one step to affect subsequent estimates, motivating joint recovery of the complete generator configuration.

\subsection{Proposed Approach and Contributions}

In this paper, we introduce \texttt{DCBA}, a model-based approach that infers the configuration of a graph generator from an observed network. Rather than directly perturbing its structure, it predicts a reusable configuration from which multiple related graphs can be sampled. Following the taxonomies~\cite{zhao2023graphdataaugmentationgraph,zhou2025data}, \texttt{DCBA} can be positioned as a learned, generation-based method for graph-level structural augmentation.

The principal contributions of this work are as follows:

\begin{itemize}[nosep, itemsep=1pt]
    \item We formulate graph augmentation through learned, amortised inversion of a stochastic graph generator, enabling multiple related graphs to be generated from a single observed network.
    \item We introduce \texttt{DCBA}, which aligns graph and configuration representations through a multi-positive contrastive objective with soft negative weighting, jointly predicts the complete generator configuration, and is instantiated using the \texttt{ABCD}~\cite{kaminski2021abcd} synthetic graph generator.
    \item We demonstrate accurate parameter recovery (mean $R^2$ of $0.761$/$0.527$ on casual/hard sets), outperforming algorithmic baselines, and show that \texttt{DCBA}-based augmentation improves community detection with \texttt{PRoCD}~\cite{qin2024pre} on real-world networks by $273\%$ in AMI on average.
\end{itemize}

\section{Problem Formulation}
\label{sec:problem}

Having introduced the initial concept of the model, this section formalises the problem without restricting it to a particular synthetic graph generator. Let

\begin{equation}
    G = (V, E, \mathbf{X})
\end{equation}

denote a graph, where $V$ is the set of vertices, $E$ is the set of edges, and $\mathbf{X} \in \mathbb{R}^{|V| \times d}$ is the matrix of vertex features, with $d$ denoting the number of features associated with each vertex.

Next, let $\mathcal{G}$ denote a stochastic graph generator parameterised by a configuration vector $\theta \in \Theta$. The generator induces a conditional distribution over graphs such that

\begin{equation}
    G \sim p_{\mathcal{G}}(G \mid \theta).
\end{equation}

Given an observed graph $G$, the objective is to learn an estimator $f_{\phi}$, parameterised by $\phi$, that predicts the underlying generator configuration:

\begin{equation}
    \widehat{\theta} = f_{\phi}(G).
\end{equation}

To address this problem, let the training dataset be defined as

\begin{equation}
    \mathcal{D}
    =
    \left\{
        \left(\theta_i, \quad G_i^{(1)}, G_i^{(2)}, \dots, G_i^{(R_i)}\right)
    \right\}_{i=1}^{N},
    \qquad
    G_i^{(r)} \sim p_{\mathcal{G}}(G \mid \theta_i),
\end{equation}

where $N$ denotes the number of distinct generator configurations and $R_i$ denotes the number of graph realisations generated for configuration $\theta_i$. Each tuple in $\mathcal{D}$ is referred to as an \emph{instance}, whereas each graph generated within an instance is referred to as a \emph{replica}.

\subsection{Model Components}

Before introducing the objective, let us note that a graph $G$ and a configuration $\theta$ constitute different modalities describing the same underlying generative process. Accordingly, we adopt a two-tower architecture inspired by CLIP-like representation learning~\cite{radford2021clip}. The two modalities are processed by separate encoders, while their representations are aligned within a shared latent space. Let

\begin{equation}
    e_{\phi_G} \colon G \mapsto h_G,
    \qquad
    e_{\phi_\theta} \colon \theta \mapsto h_\theta
\end{equation}

denote the graph encoder, parameterised by $\phi_G$, and the configuration encoder, parameterised by $\phi_\theta$, respectively. Both encoders produce representations in the same $k$-dimensional latent space:

\begin{equation}
    h_{G,i}^{(r)} = e_{\phi_G}\left(G_i^{(r)}\right) \in \mathbb{R}^{k},
    \qquad
    h_{\theta,i} = e_{\phi_\theta}\left(\theta_i\right) \in \mathbb{R}^{k}.
\end{equation}

The configuration encoder forms part of an autoencoder whose decoder, parameterised by $\phi_D$, is defined as

\begin{equation}
    d_{\phi_D} \colon \mathbb{R}^{k} \rightarrow \Theta.
\end{equation}

The decoder is shared between the configuration reconstruction and graph-to-configuration paths:

\begin{equation}
    \widetilde{\theta}_i = d_{\phi_D}\left(h_{\theta,i}\right),
    \qquad
    \widehat{\theta}_i^{(r)} = d_{\phi_D}\left(h_{G,i}^{(r)}\right).
\end{equation}

Consequently, the estimator introduced previously can be expressed as the composition

\begin{equation}
    f_\phi = d_{\phi_D} \circ e_{\phi_G},
    \qquad
    \phi = \left\{ \phi_G, \phi_\theta, \phi_D \right\}.
\end{equation}

The autoencoding path encourages the configuration tower to learn a stable and informative representation of the generator parameter space. The contrastive objective subsequently aligns graph representations with this space, allowing $d_{\phi_D}$ to be applied directly to graph representations at inference time. Although $e_{\phi_\theta}$ participates in the training procedure, the graph-to-configuration inference path depends only on $e_{\phi_G}$ and the shared $d_{\phi_D}$.

\subsection{Configuration Prediction and Reconstruction}
\label{sec:problem:baseloss}

Let $\mathcal{I}$ denote the set of generator configurations included in a mini-batch, and let $\mathcal{R}_i$ denote the set of replica indices sampled for configuration $\theta_i$. The graph-to-configuration regression objective is

\begin{equation}
    \label{eq:loss_reg}
    \mathcal{L}_{\mathrm{reg}} =
    \frac{1}{\sum_{i \in \mathcal{I}}|\mathcal{R}_i|}
    \sum_{i \in \mathcal{I}}
    \sum_{r \in \mathcal{R}_i}
    \ell_{\Theta}\left(\widehat{\theta}_i^{(r)},\theta_i\right),
\end{equation}

whereas the reconstruction objective of the configuration autoencoder is defined as

\begin{equation}
    \label{eq:loss_rec}
    \mathcal{L}_{\mathrm{rec}} =
    \frac{1}{|\mathcal{I}|}
    \sum_{i \in \mathcal{I}}
    \ell_{\Theta}\left(\widetilde{\theta}_i,\theta_i\right).
\end{equation}

Here, $\ell_{\Theta} \colon \Theta \times \Theta \rightarrow \mathbb{R}_+$ denotes a discrepancy function defined over pairs of generator configurations. Its particular form may depend on the parameterisation of the considered generator $\mathcal{G}$. In the simplest case, it may be instantiated as the mean squared error, whereas generator-specific implementations may additionally account for differences in parameter scales or penalise predicted configurations that violate the feasibility constraints of the generator, such as non-negativity.

\subsection{Multi-positive Contrastive Objective with Soft Negative Weighting}
\label{sec:problem:supcon}

As $\mathcal{G}$ is nondeterministic, several graphs may be generated from the same configuration. These replicas should occupy nearby regions of the latent space despite the differences introduced by the stochastic generation process. Thus, we employ a multi-positive supervised contrastive objective~\cite{khosla2020supervised} with soft negative weighting, extending the bidirectional alignment principle used in~\cite{radford2021clip}.

The normalised latent representations are defined as

\begin{equation}
    z_{G,i}^{(r)} = \frac{h_{G,i}^{(r)}}{\left\|h_{G,i}^{(r)}\right\|_2}, \qquad z_{\theta,i} = \frac{h_{\theta,i}}{\left\|h_{\theta,i}\right\|_2}.
\end{equation}

For a given mini-batch, let

\begin{equation}
    \mathcal{Z}_G = \left\{z_{G,i}^{(r)} \colon i \in \mathcal{I}, \; r \in \mathcal{R}_i\right\}, \qquad \mathcal{Z}_\theta = \left\{z_{\theta,i} \colon i \in \mathcal{I}\right\}
\end{equation}

denote the collection of graph representations and the collection containing one configuration representation for each generator configuration, respectively. The complete representation pool is

\begin{equation}
    \mathcal{Z} = \mathcal{Z}_G \cup \mathcal{Z}_\theta.
\end{equation}

Each representation $z \in \mathcal{Z}$ is associated with an instance label $y(z)$ and its corresponding generator configuration $\theta(z)$. For an anchor $a \in \mathcal{Z}$, the positive and negative sets are defined, respectively, as

\begin{equation}
    \mathcal{P}(a) = \left\{z \in \mathcal{Z}\setminus\{a\} \colon y(z)=y(a)\right\}, \qquad \mathcal{N}(a) = \left\{z \in \mathcal{Z} \colon y(z)\neq y(a)\right\}.
\end{equation}

Thus, for a graph anchor $z_{G,i}^{(r)}$, the positive set contains the configuration representation $z_{\theta,i}$ and all other graph replicas generated from $\theta_i$. For a configuration anchor $z_{\theta,i}$, the positive set contains all graph replicas generated from that configuration.

Since all representations are normalised, their similarity scaled by the $\tau_s>0$ constant is defined as

\begin{equation}
    s(a,z) = \frac{a^\mathsf{T}z}{\tau_s}.
\end{equation}

Generator configurations form a continuous parameter space. Consequently, two configurations located close to one another should not be treated as equally strong negatives as configurations lying far apart. For a negative pair $(a,z)$, we therefore define the distance-dependent weight

\begin{equation}
    w(a,z) = 1 - \exp\left(-\frac{\left\|\theta(a)-\theta(z)\right\|_2}{\tau_{w}}\right),
\end{equation}

where $\tau_{w}>0$ controls the rate at which the weight increases with configuration distance. The weight approaches zero for nearly identical configurations and approaches one as the distance between configurations increases.

Therefore, for an anchor $a$, the normalising term is defined as

\begin{equation}
    D(a) =
    \sum_{p \in \mathcal{P}(a)} \exp\left(s(a,p)\right) +
    \sum_{n \in \mathcal{N}(a)} w(a,n)\exp\left(s(a,n)\right).
\end{equation}

The term $D(a)$ can be interpreted as the total similarity mass of all representations competing with respect to anchor $a$. The more similar a representation (whether positive or negative) is to the anchor, the more strongly it contributes to $D(a)$. For negatives, this contribution is additionally modulated by $w(a,n)$. Consequently, $D(a)$ ensures that the similarity of each positive representation is evaluated relative to the remaining representations in the mini-batch.

Finally, for a collection of anchors $\mathcal{A}\subseteq\mathcal{Z}$, the multi-positive contrastive loss, averaged over all anchor--positive pairs, is defined as

\begin{equation}
    \mathcal{L}_{\mathrm{MP}}(\mathcal{A}) =
    -\frac{1}{\sum_{a \in \mathcal{A}}\left|\mathcal{P}(a)\right|}
    \sum_{a \in \mathcal{A}}
    \sum_{p \in \mathcal{P}(a)}
    \log\frac{\exp\left(s(a,p)\right)}{D(a)}.
\end{equation}

The loss is evaluated bidirectionally. Graph representations act as anchors in the first direction, whereas configuration representations act as anchors in the second direction:

\begin{equation}
    \mathcal{L}_{\mathrm{con}} = \mathcal{L}_{\mathrm{MP}}\left(\mathcal{Z}_G\right) + \mathcal{L}_{\mathrm{MP}}\left(\mathcal{Z}_\theta\right).
\end{equation}

In conclusions, the objective encourages graph replicas generated from the same configuration to form a coherent cluster around their configuration representation, while configurations that are distant in parameter space provide a stronger repulsive signal than nearby configurations.

\subsection{Joint Learning Objective}

The complete model is trained by minimising

\begin{equation}
    \phi^{\star} = \arg\min_{\phi} \mathcal{L}_{\mathrm{total}} \left( \phi; \mathcal{D}\right),
\end{equation}

where

\begin{equation}
    \label{eq:total_loss_theory}
    \mathcal{L}_{\mathrm{total}} =
    \lambda_{\mathrm{reg}} \mathcal{L}_{\mathrm{reg}} +
    \lambda_{\mathrm{rec}} \mathcal{L}_{\mathrm{rec}} +
    \lambda_{\mathrm{con}} \mathcal{L}_{\mathrm{con}}.
\end{equation}

The regression term grounds the latent representation in the original generator-parameter space, the reconstruction term regularises the configuration tower, and the contrastive term aligns graph and configuration representations while promoting invariance to the stochasticity of the graph generator.

\section{Framework Instantiation -- \texttt{DCBA}}
\label{sec:method}

The framework introduced in Sec.~\ref{sec:problem} is instantiated using the \texttt{ABCD} model~\cite{kaminski2021abcd}. Before discussing the \texttt{DCBA} architecture (shown in Fig.~\ref{fig:architecture}), we first describe the generator itself and the dataset constructed from its realisations.

\subsection{Employed Graph Generator}
\label{sec:method:abcd}

\texttt{ABCD} is a graph generator attractive due to its rich modelling space tailored for social networks and the non-trivial retrieval of input parameters from its realisations. The model is \emph{scale-free} -- both the degree sequence and the community-size sequence are drawn from power law distributions to mimic the heavy-tailed characteristic of real-world networks. Moreover, each generated graph is accompanied by a ground-truth community assignment.

Succinctly, \texttt{ABCD} generates a graph with $n$ vertices as the union of independently sampled components: a background graph carrying inter-community noise and one graph associated with each community. A degree sequence $\mathbf{d}$ and a community-size sequence $\mathbf{m}$ are sampled from truncated power-law distributions. For each vertex $v$, its prescribed degree $d_v$ is divided into a community contribution $d_v^{\mathrm{com}} \approx (1-\xi)d_v$ and a background contribution $d_v^{\mathrm{bg}} \approx \xi d_v$. Hence, the mixing parameter $\xi \in [0,1]$ controls the expected proportion of a vertex's edges assigned to the background graph. Values approaching zero produce strongly separated communities, whereas values approaching one progressively weaken the community structure. The independently generated components are subsequently combined and simplified by removing self-loops and multiple edges.

\texttt{ABCD+o}~\cite{kaminski2023abcdo} extends this construction with a specified number of outlier vertices -- $n_{\text{out}}$. Outliers are not assigned to any community graph, and their incident edges are generated exclusively within the background component. Hence, they have no dominant community affiliation. Setting $n_{\text{out}}$ to zero recovers the original \texttt{ABCD} model. We therefore adopt \texttt{ABCD+o} as the generator underlying \texttt{DCBA}, since it subsumes the original model while introducing an additional characteristic to be recovered.

\subsection{Utilised Dataset}
\label{sec:method:dataset}

Training data were produced by sampling 10\,000 candidate \texttt{ABCD+o} configurations $\theta$ from the ranges reported in Tab.~\ref{tab:abcdo-params} and generating independent stochastic replicas for each accepted configuration. Concretely, for every candidate configuration, each parameter was sampled independently and uniformly from its corresponding interval. Configurations that violated the feasibility constraints of \texttt{ABCD+o} were discarded. For details on the dataset generation, see App.~\ref{app:dataset-generation}.

The procedure yielded 8,744 accepted configurations, each constituting one \emph{instance} in the sense defined in Sec.~\ref{sec:problem}. For every instance, ten independent graph \emph{replicas} were generated using distinct random seeds, resulting in a dataset of 87,440 graphs together with their community assignments. Separately, a boundary challenge set comprising 486 configurations and 4,860 graphs was generated from combinations of parameter values located at the limits of the ranges reported in Tab.~\ref{tab:abcdo-params}. It was intended to evaluate the model on challenging configurations and thereby assess its robustness.

\vspace{-1em}
\begin{table}[ht]
    \centering
    \caption{Sampling ranges used to construct the dataset. Graph-size-dependent parameters are drawn as fractions of $n$ and subsequently converted to valid integer-valued parameters of \texttt{ABCD+o}.}
    \label{tab:abcdo-params}
    \small
    \begin{tabularx}{\textwidth}{@{}l l X@{}}
        \toprule
        Symbol & Sampling range & Description \\
        \midrule
        $n$ & $[1\,000,\ 10\,000]$ & Number of nodes in the graph. \\
        $t_1$ & $[2.0,\ 3.0]$ & Power-law exponent of the degree distribution. \\
        $d_{\min}$ & $[10^{-10},\ 0.005]\cdot n$ & Minimum node degree. \\
        $d_{\max}$ & $[0.005,\ 0.5]\cdot n$ & Maximum node degree. \\
        $t_2$ & $[1.0,\ 2.0]$ & Power-law exponent of the community-size distribution. \\
        $c_{\min}$ & $[10^{-10},\ 0.05]\cdot n$ & Minimum community size. \\
        $c_{\max}$ & $[0.05,\ 0.5]\cdot n$ & Maximum community size. \\
        $\xi$ & $[0.01,\ 1.0]$ & Fraction of edges placed in the background graph. \\
        $n_{\text{out}}$ & $[0.0,\ 0.05]\cdot n$ & Number of outlier nodes (\texttt{ABCD+o}) excluded from every community. \\
        \bottomrule
    \end{tabularx}
\end{table}

\subsection{Proposed Model}
\label{sec:model}

\begin{figure}[ht]
    \centering
    \includegraphics[width=\linewidth]{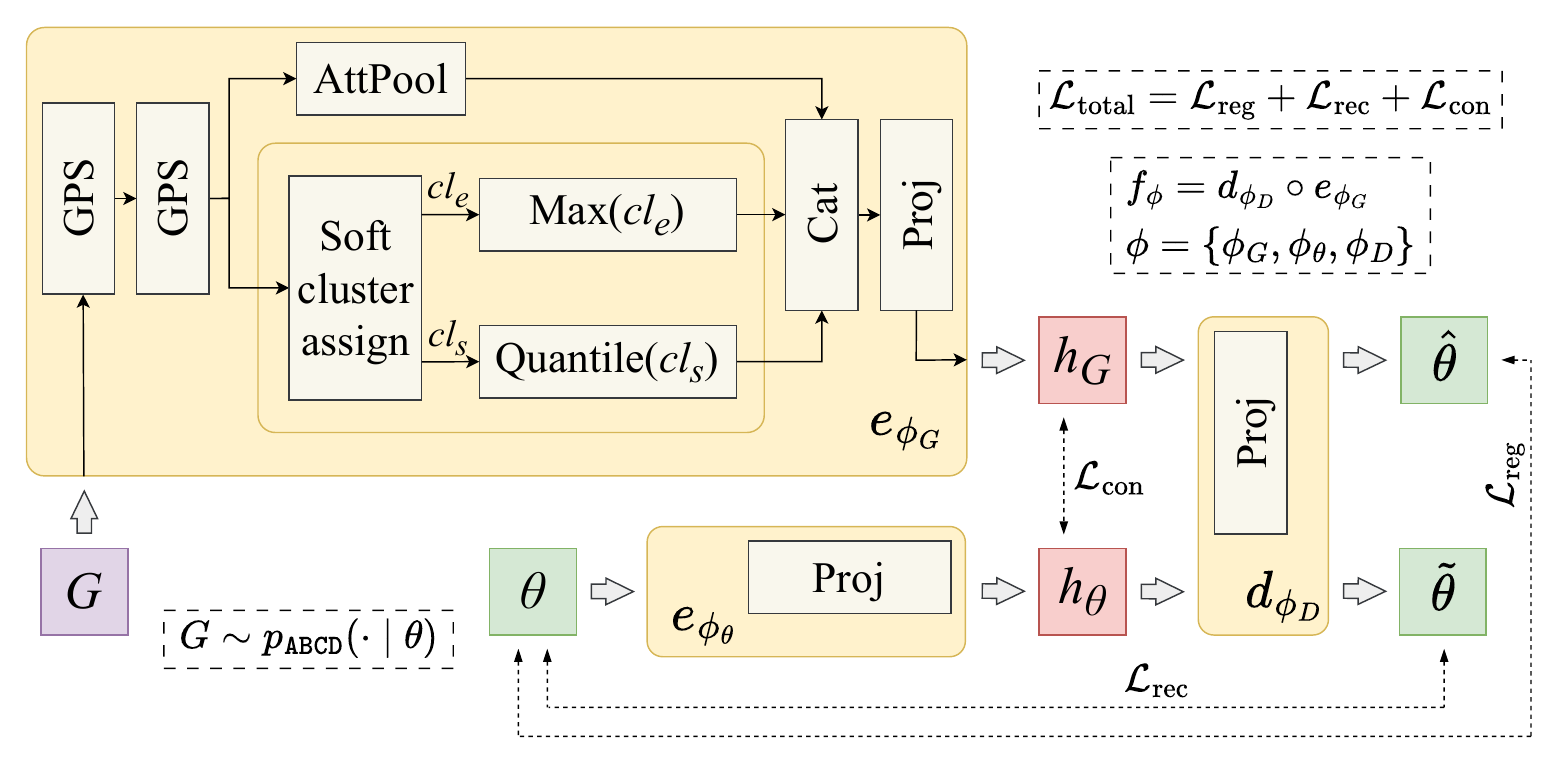}
    \caption{
        Architecture of \texttt{DCBA}. During training, \texttt{ABCD} configurations $\theta$ and their graph realisations $G$ are mapped by $e_{\phi_\theta}$ and $e_{\phi_G}$, respectively, into a shared latent space. Representations originating from the same configuration are pulled together, whereas those of different configurations are pushed apart with a strength proportional to their distance in the configuration space. A soft-clustering pool assigns vertices to $K$ latent slots and summarises their embeddings $\mathit{cl}_e$ and sizes $\mathit{cl}_s$ independently of slot order. At inference, only the path $f_\phi=d_{\phi_D}\circ e_{\phi_G}$ is used to recover $\hat{\theta}$ from an observed graph.
    }
    \label{fig:architecture}
\end{figure}

Everything the model can infer about an \texttt{ABCD} realisation has to fit into $h_G$. The central design question is therefore how much of a graph's macrostructure can be compressed into a single vector. In particular, $h_G$ must simultaneously capture the degree distribution, community organisation, and the extent of inter-community mixing. Since the components of $\theta$ are predicted jointly, information omitted by the graph encoder cannot be recovered further down the model. Figure~\ref{fig:architecture} shows the architecture designed around this requirement, while App.~\ref{app:loss} describes the employed loss function.

\vspace{-1em}

\paragraph{Encoding Vertices} The graph encoder $e_{\phi_G}$ is composed of GraphGPS layers~\cite{rampavsek2022recipe}. Each layer combines local GIN message passing~\cite{xu2018powerful} with global multi-head self-attention, implemented using a linear approximation~\cite{choromanski2021rethinking} to reduce memory requirements for the graph sizes in Tab.~\ref{tab:abcdo-params}. This allows each layer to combine local and graph-wide information. Depending on the experimental setting, vertices receive either a constant feature or features encoding community membership.

\paragraph{Aggregating Vertices into $h_G$} Vertex embeddings are aggregated along two branches. Attention pooling~\cite{li2019graph} provides a weighted average, while the second branch softly assigns vertices to $K$ learned slots and reads out order statistics of their representations and sizes. Consequently, $h_G$ captures both the averaged vertex representation and how vertex embeddings accumulate within slots.

\paragraph{Soft-clustering Regularisation} To make the slot sizes behave as counts rather than diffuse mass, the training objective includes an additional entropy term that encourages each vertex to concentrate its assignment on a single slot. The slots need not correspond to communities, and $K$ only bounds how many may be used. The complete formulation is provided in App.~\ref{app:aux-cluster}.

\paragraph{Representing Configurations} The configuration encoder $e_{\phi_\theta}$ and the shared decoder $d_{\phi_D}$ are multilayer perceptrons with mirrored layer widths. As several components of $\theta$ scale with the graph order, both $\ell_\Theta$ and the weight $w(a,z)$ would otherwise be dominated by $n$. Therefore, configurations are scaled before being used by the model, with size-related components transformed logarithmically and the remaining ones linearly (App.~\ref{app:loss:scaler}). The decoder outputs lie in the same normalised space.

\subsection{Baseline Inverse Model}
\label{sec:method:baseline}

As a non-learned reference point for \texttt{DCBA}, we developed an algorithmic baseline that estimates the vector $\theta$ directly from an observed \texttt{ABCD+o} graph. Its formulation is provided in App.~\ref{app:baseline}. The baseline follows a recovery procedure in which the components of $\theta$ are estimated independently, similarly to~\cite{brodka2026twinning}. Parameters directly observable in the graph, such as the minimum degree, are obtained therefrom, whereas the remaining parameters are estimated from the corresponding structural characteristics. We consider two variants. \texttt{Base-T} uses the ground-truth community assignment associated with the generated \texttt{ABCD+o} graph, whereas \texttt{Base-L} detects communities using the Leiden algorithm~\cite{traag2019leiden}. Thus, it can be applied when the true community assignment is unavailable.

\section{Experiments}
\label{sec:experiments}

The trained model was compared with the algorithmic baseline using the coefficient of determination, $R^2$, computed separately for each component of $\theta$ between the ground-truth and predicted values. To prevent data leakage, all graphs generated from a given $\theta$ were assigned jointly to the same data fold.


\subsection{Prediction Performance}

We compare three learned variants: the standard \texttt{DCBA}; \texttt{DCBA-Comm}, which uses the size of the community containing each vertex as a node feature; and \texttt{DCBA-Comm64}, additionally trained with a batch size of $64$. As \texttt{Base-T} uses ground-truth community information, it should be compared to \texttt{DCBA-Comm} and \texttt{DCBA-Comm64}, whereas \texttt{Base-L} corresponds to \texttt{DCBA}. Let us recall that the baselines also have an inherent advantage for directly observable parameters such as $n$ and $d_{\min}$.

\vspace{-1em}
\begin{table}[ht!]
    \centering
    \caption{Parameter-wise $R^2$ on the casual and hard test sets; first- and second-best results within each split are shown in dark and pale green, respectively; the bottom row reports the overall mean.}
    \label{tab:r2-results}
    \addtolength{\tabcolsep}{-0.03em}
    \small
    \begin{tabular}{@{}l*{10}{r}@{}}
        \toprule
        & \multicolumn{5}{c}{Casual test set}
        & \multicolumn{5}{c}{Hard test set} \\
        \cmidrule(lr){2-6}
        \cmidrule(l){7-11}

        Param.
        & \texttt{Base-T}
        & \texttt{Base-L}
        & \texttt{DCBA}
        & \makecell{\texttt{DCBA}\\\texttt{Comm}}
        & \makecell{\texttt{DCBA}\\\texttt{Comm64}}
        & \texttt{Base-T}
        & \texttt{Base-L}
        & \texttt{DCBA}
        & \makecell{\texttt{DCBA}\\\texttt{Comm}}
        & \makecell{\texttt{DCBA}\\\texttt{Comm64}} \\
        \midrule

        $n$
        & \first{1.000}
        & \first{1.000}
        & \second{0.964}
        & 0.948
        & 0.951
        & \first{1.000}
        & \first{1.000}
        & \second{0.930}
        & 0.808
        & 0.704 \\

        $t_1$
        & -5.425
        & -5.425
        & \first{0.959}
        & 0.914
        & \second{0.927}
        & -66.760
        & -66.760
        & \first{0.666}
        & 0.489
        & \second{0.606} \\

        $d_{\min}$
        & \first{1.000}
        & \first{1.000}
        & \second{0.971}
        & 0.946
        & 0.970
        & \first{1.000}
        & \first{1.000}
        & \second{0.946}
        & 0.791
        & 0.731 \\

        $d_{\max}$
        & 0.661
        & 0.661
        & \first{0.801}
        & 0.740
        & \second{0.779}
        & 0.449
        & 0.449
        & \second{0.590}
        & \first{0.680}
        & 0.470 \\

        $t_2$
        & -2.031
        & -16.130
        & 0.007
        & \first{0.105}
        & \second{0.089}
        & -86.160
        & -36.430
        & \second{-0.012}
        & -0.082
        & \first{0.204} \\

        $c_{\min}$
        & 0.635
        & -0.048
        & 0.439
        & \second{0.861}
        & \first{0.876}
        & \first{0.966}
        & -0.194
        & -0.013
        & \second{0.778}
        & 0.639 \\

        $c_{\max}$
        & 0.619
        & 0.404
        & 0.544
        & \second{0.782}
        & \first{0.790}
        & \second{0.471}
        & -0.024
        & 0.328
        & \first{0.615}
        & 0.446 \\

        $\xi$
        & 0.923
        & 0.637
        & 0.859
        & \first{0.981}
        & \second{0.980}
        & \first{0.770}
        & 0.178
        & 0.040
        & 0.471
        & \second{0.548} \\

        $n_{\text{out}}$
        & -2.760
        & -1.512
        & 0.306
        & \second{0.480}
        & \first{0.484}
        & -45.250
        & -0.356
        & \first{0.214}
        & \second{0.197}
        & 0.069 \\

        \midrule
        $\overline{R^2}$
        & -0.598
        & -2.157
        & 0.650
        & \second{0.751}
        & \first{0.761}
        & -21.501
        & -11.237
        & 0.410
        & \first{0.527}
        & \second{0.491} \\
        \bottomrule
    \end{tabular}
\end{table}

Table~\ref{tab:r2-results} presents results for both sets. On the ``casual'' one, all \texttt{DCBA} variants substantially outperform the baselines for the power-law exponents and $n_{\mathrm{out}}$. Community features markedly improve the recovery of $c_{\min}$, $c_{\max}$, and $\xi$, with \texttt{DCBA-Comm64} achieving the highest overall $\overline{R^2}$. On the hard set, the algorithmic baselines deteriorate sharply for $t_1$, $t_2$, and $n_{\mathrm{out}}$, whereas the learned variants remain substantially more robust. Here, \texttt{DCBA-Comm} attains the highest overall score, while the vanilla model performs best for several parameters, indicating better generalisation under distribution shift than the larger-batch variant. Community features remain particularly beneficial for community-related parameters. Finally, $t_2$ remains the most difficult parameter to recover, as its estimation is based on fewer community sizes (less than $\sqrt{n}$) than the sequence of node degrees ($n$) governed by $t_1$. Outliers are also structurally similar to noise and hence challenging for both the baselines and \texttt{DCBA}.

\vspace{-0.25em}
\subsection{Applicability to the Community Detection Task}

Since the framework was instantiated with \texttt{ABCD}, a model designed primarily for benchmarking community detection methods~\cite{kaminski2021abcd}, we evaluated whether \texttt{DCBA}-based augmentation can improve a trainable community detection model. To this end, we selected \texttt{PRoCD}~\cite{qin2024pre}, a neural method that does not require prior knowledge of the number of communities and can be trained on synthetic corpora, making it well suited to data augmentation. The experiment comprised two complementary parts.

The first one used a controlled synthetic setting. Since \texttt{PRoCD}'s robustness deteriorated for networks with weak communities, creating room for augmentation, we selected $22$ graphs with $\xi > 0.25$ from the held-out ``casual'' dataset (Sec.~\ref{sec:method:dataset}). Performance without adaptation constituted the baseline condition (\texttt{B}). For each target, we then fine-tuned (thrice with different random seeds) the model on cohorts of $20$ synthetic \texttt{ABCD} graphs generated under three conditions. In the random condition (\texttt{R}), configurations were sampled independently of the target, controlling for the general effect of \texttt{ABCD} data. In the \texttt{DCBA} condition (\texttt{D}), the cohort was generated from the configuration inferred by \texttt{DCBA} trained without ground-truth community information. Finally, in the oracle condition (\texttt{O}), graphs were generated from the configuration used to create the target graph. Thus, \texttt{R} controls for non-targeted augmentation, while \texttt{O} provides a reference with exact knowledge of the target.

\vspace{-1em}
\begin{table}[ht]
    \centering
    \caption{Community detection with \texttt{PRoCD} on unseen \texttt{ABCD} graphs under different fine-tuning regimes. AMI is averaged while $\Delta$\texttt{B} and $p$ compare each regime with \texttt{B} using a paired Wilcoxon test.}
    \label{tab:dcba-augm-synth}
    \small
    \begin{tabular}{llccc}
    \toprule
    Mode & Fine-tuning corpus & AMI & $\Delta$\texttt{B} & $p$ \\
    \midrule
    \texttt{B} & none (frozen checkpoint) & $0.189$ & -- & -- \\
    \texttt{R} & random \texttt{ABCD} graphs & $0.378$ & $+0.189$ & $0.0057$ \\
    \texttt{D} & inferred $\hat{\theta}$ with \texttt{DCBA} & $\mathbf{0.493}$ & $\mathbf{+0.304}$ & $\mathbf{<10^{-4}}$ \\
    \texttt{O} & ground-true $\theta$ & $0.541$ & $+0.351$ & $<10^{-4}$ \\
    \bottomrule
    \end{tabular}
\end{table}

The results in Tab.~\ref{tab:dcba-augm-synth} support the applicability of the proposed augmentation method. Fine-tuning on cohorts generated from \texttt{DCBA}-inferred configurations improves AMI (Adjusted Mutual Information~\cite{vinh2009information}, a permutation-invariant clustering score adjusted for chance, with $1$ denoting perfect agreement and $0$ random expectation) by $0.304$ over the frozen checkpoint, with improvements observed on 20 of 22 target graphs ($p<10^{-4}$). Importantly, it also outperforms non-targeted \texttt{ABCD} augmentation by $0.115$ on average (20 of 22 targets, $p<10^{-3}$), indicating additional benefit from target matching. Moreover, \texttt{DCBA}-based augmentation recovers $88\%$ of the oracle improvement. 

Having demonstrated the utility of \texttt{DCBA} in augmenting in-distribution graphs, we next considered a real-world setting. The experiment was repeated on six networks with ground-truth communities (see Tab.~\ref{tab:dcba-augm-real}). For each target, we evaluated the published \texttt{PRoCD} checkpoint directly and after fine-tuning, repeated with three random seeds, on 20 \texttt{ABCD} graphs generated from configurations inferred by either \texttt{Base-L} or vanilla \texttt{DCBA}. The results further support the applicability of \texttt{DCBA}-based augmentation, although the gains are less uniform than in the synthetic setting. Across the six networks, \texttt{DCBA} achieves a mean AMI of $0.271$, compared with $0.209$ for \texttt{Base-L} and $0.175$ for the default checkpoint, corresponding to improvements of $29\%$ and $55\%$, respectively. The largest gain is observed for PolBlogs, while EuroSiS and Cora-Full show no improvement over the original checkpoint, indicating that target-specific augmentation is most beneficial when there is sufficient room for adaptation.

\vspace{-1em}
\begin{table}[ht]
    \centering
    \caption{Community detection with \texttt{PRoCD} on six real-world networks from~\cite{ivashkin2016logarithmic} under different fine-tuning regimes, measured by average AMI against ground-truth communities with standard deviation over three runs for the best-performing regime.}
    \label{tab:dcba-augm-real}
    \small
    \begin{tabular}{lcccc}
        \toprule
        Graph & Pretrained & \texttt{Base-L} & \texttt{DCBA} & $\sigma$ \\
        \midrule
        PolBlogs~\cite{adamic2005political} & $0.021$ & $0.020$ & $\mathbf{0.291}$ & $0.000$ \\
        Newsgroup 5cl-1~\cite{yen2007graph} & $0.018$ & $0.025$ & $\mathbf{0.031}$ & $0.000$ \\
        Newsgroup 5cl-2~\cite{yen2007graph} & $0.112$ & $0.194$ & $\mathbf{0.263}$ & $0.053$ \\
        Newsgroup 5cl-3~\cite{yen2007graph} & $0.096$ & $0.226$ & $\mathbf{0.236}$ & $0.001$ \\
        EuroSiS~\cite{eurosis} & $\mathbf{0.603}$ & $0.602$ & $\mathbf{0.603}$ & $0.000$ \\
        Cora-Full: OS~\cite{vsubelj2013model} & $\mathbf{0.199}$ & $0.187$ & $\mathbf{0.199}$ & $0.000$ \\
        \bottomrule
    \end{tabular}
\end{table}

\section{Conclusions}
\label{sec:conclusions}

This paper proposes a new approach to graph data augmentation through inverse graph modelling rather than micro-level perturbations. Inspired by \texttt{CLIP}~\cite{radford2021clip}, the introduced architecture aligns graph and generator-configuration representations through contrastive learning and uses a configuration autoencoder to support stable training. Experiments show that its \texttt{ABCD}-specific instantiation, \texttt{DCBA}, robustly recovers most generator parameters and outperforms the algorithmic baseline, except for directly observable quantities. The study also validates its utility in both curated and real-world community detection settings, showing that fine-tuning corpora obtained with the graph--\texttt{DCBA}--\texttt{ABCD} pipeline improve \texttt{PRoCD} by up to $161\%$ and $273\%$ in AMI, respectively. However, the major limitation is its dependence on a particular generator, whose assumptions may not necessarily hold for an observed graph. Future work should therefore address generator selection as a preceding stage. 

\vfill

\paragraph{Acknowledgements} {\footnotesize This research was partially supported by: Horizon Europe, EU, grant no. 10108632; National Science Centre, Poland, grant no. 2022/45/B/ST6/04145; Polish National Agency for Academic Exchange, Strategic Partnerships, grant no. BPI/PST/2024/1/00129/U/00001; Wroc{\l}aw University of Science and Technology, Academia Professorum Iuniorum and  Minigrants projects. Views and opinions expressed here are those of the authors and do not necessarily reflect those of the funding agencies.}

\printbibliography

\appendix

\section{\texttt{ABCD+o}-specific Loss}
\label{app:loss}

Following the general formulation of the training objective from Sec.~\ref{sec:problem}, we show below the concrete loss function used to train \texttt{DCBA}, as described in Sec.~\ref{sec:method}. As presented in Eq.~\ref{eq:total_loss_theory}, the objective comprises three principal terms addressing different aspects of the problem. In \texttt{DCBA}, their corresponding weights were set, based on empirical experience and intuition, to $\lambda_{reg}=1.0$, $\lambda_{rec}=1.0$, and $\lambda_{con}=0.3$. The contrastive term retains the form defined in Sec.~\ref{sec:problem:supcon}, with its two normalisation parameters set to $\tau_s=0.07$ and $\tau_w=1.0$. The regression and reconstruction terms, only outlined in the main text, are instantiated using an \texttt{ABCD+o}-specific configuration loss (see App.~\ref{app:loss:configuration}). In addition, the cluster-pooling branch of the graph encoder (Sec.~\ref{sec:model}) contributes an auxiliary clustering objective, which is detailed below. The following text also describes the normalisation of the configuration space on which these terms operate.

\subsection{Configuration Scaling}
\label{app:loss:scaler}

The components of $\theta$ span incomparable scales. The two power-law exponents and the mixing parameter are of order one, whereas $n$ reaches $10^4$, as do the components defined as fractions of the graph order. Consequently, when computed on raw values, both a squared-error loss over $\Theta$ and the Euclidean distance between two configurations are dominated by $n$ and the components that scale with it. Each configuration is therefore mapped to normalised coordinates using a scaler $\sigma$ before being passed to the model, and both model outputs are produced in these coordinates. The setup reported in Sec.~\ref{sec:experiments} uses a scaler that compresses the size-related components logarithmically and maps the remaining components linearly to their admissible ranges, as defined in Eq.~\ref{eq:log_scaler}:

\begin{equation}
    \label{eq:log_scaler}
    \sigma(\theta)_j =
    \begin{cases}
        \log(1+\theta_j) \,/\, \log(1+n_{\max}) & j = n, \\
        \left(\theta_j - 1\right) \,/\, \left(t_{\max}-1\right) & j \in \{t_1, t_2\}, \\
        \theta_j & j = \xi, \\
        \log(1+\theta_j) \,/\, \log(1+n) & j \in \{c_{\max}, d_{\max}, n_{\text{out}}\}, \\
        \log(1+\theta_j) \,/\, \log(1+c_{\max}) & j = c_{\min}, \\
        \log(1+\theta_j) \,/\, \log(1+d_{\max}) & j = d_{\min}
    \end{cases}
\end{equation}

Here, the components of $\theta$ are indexed by parameter name, with $n_{\max}=10^4$ and $t_{\max}=5$. Each logarithmic coordinate is divided by its anchor, the quantity the generator forbids the component to exceed, clamped from below by one so that no denominator vanishes. The bounds are fixed in advance rather than estimated from the training split, $t_1,t_2\geq1$ and $\xi\in[0,1]$ being constraints of \texttt{ABCD+o} and $n_{\max}$, $t_{\max}$ chosen ceilings exceeding the ranges of Tab.~\ref{tab:abcdo-params}.

It is important to note that the map is invertible. Within the ordering penalty of the following subsection the inverse is applied as written, so that the penalty remains differentiable with respect to the model output. A configuration handed back to the generator has $n$, $c_{\min}$, $c_{\max}$, $d_{\min}$, $d_{\max}$, and $n_{\text{out}}$ rounded to integers. Alternative scalers are compared in App.~\ref{app:ablation}.

\subsection{Constraint-aware Configuration Loss}
\label{app:loss:configuration}

Let us specify the instantiation of the configuration-space discrepancy function $\ell_{\Theta}$ (Sec.~\ref{sec:problem:baseloss}) used in \texttt{DCBA}. The loss combines a parameter-wise weighted squared error with soft penalties discouraging predictions that violate the admissibility constraints of the generator.

The nine-dimensional configuration of \texttt{ABCD+o} can be written in the order used in Tab.~\ref{tab:abcdo-params} as

\begin{equation}
    \theta =
    \left(n,t_1,d_{\min},d_{\max},t_2,c_{\min},c_{\max},\xi,n_{\text{out}}\right) \in \Theta.
\end{equation}

Before being passed to the model, each configuration vector is transformed by the scaler of App.~\ref{app:loss:scaler}. Both the graph-derived prediction $\hat{\theta}$ and the reconstruction $\tilde{\theta}$ are produced in this scaled parameter space. For compactness, let $\theta'$ denote either of these model outputs. The discrepancy between $\theta'$ and the corresponding target $\theta$ is evaluated in the scaled space, whereas the generator-specific constraints are evaluated after both vectors have been transformed back to their original parameter scale.

The parameter-wise discrepancy is measured using a weighted mean squared error,

\begin{equation}
    \ell_{\mathrm{mse}}(\theta',\theta) =
    \frac{1}{q} \sum_{j=1}^{q} \omega_j\left(\theta'_j-\theta_j\right)^2,
\end{equation}

where $q=9$ and $\omega_j \geq 0$ denotes the weight assigned to the $j$-th configuration component. Setting $\omega_j = 1$ for every component recovers the ordinary mean squared error. The parameter weights are treated as hyperparameters, and the settings considered in the experiments are reported in App.~\ref{app:ablation}.

A range penalty discourages outputs from lying outside the expected range of the scaled parameter space. For scalers mapping the configuration components to normalised coordinates, it is defined as

\begin{equation}
    \ell_{\text{range}}(\theta') = \sum_{j=1}^{q}\left[-\theta'_j\right]_{+}^{2} + \sum_{j=2}^{q}\left[\theta'_j-1\right]_{+}^{2},
\end{equation}

where $[x]_{+}=\max(0,x)$. The first term penalises negative values of all configuration components, whereas the second penalises values exceeding the upper normalisation bound. The graph-size parameter $n$ is exempt from the upper-range penalty, allowing the estimator to predict graphs larger than those represented in the range assumed by the scaler. When no normalisation is applied, the second sum is omitted, since raw parameter values may naturally exceed one.

The ordering penalty enforces relationships between the \texttt{ABCD+o} parameters in their original scale. To avoid introducing additional notation, in this context $\theta'$ denotes the model output after reversing the scaling transformation. Each constraint violation is divided by the ground-truth graph order $n$, making it dimensionless and reducing its dependence on graph size. The penalty is defined as

\begin{align}
    \ell_{\text{ord}}(\theta',\theta) = {}&
    \left[ \frac{\theta'_{c_{\min}} - \theta'_{c_{\max}}}{n} \right]_{+}^{2} +
    \left[ \frac{\theta'_{d_{\min}} - \theta'_{d_{\max}}}{n} \right]_{+}^{2} \nonumber \\ &+ 
    \left[ \frac{\theta'_{c_{\max}} - \theta'_{n}}{n} \right]_{+}^{2} +
    \left[ \frac{\theta'_{d_{\max}} - \theta'_{n}}{n} \right]_{+}^{2} +
    \left[ \frac{\theta'_{n_{\text{out}}}-\theta'_{n}}{n} \right]_{+}^{2}.
\end{align}

The individual terms penalise violations of

\begin{equation}
    c_{\min}\leq c_{\max}, \qquad d_{\min}\leq d_{\max}, \qquad c_{\max}\leq n, \qquad d_{\max}\leq n, \qquad n_{\text{out}}\leq n.
\end{equation}

Because the positive-part operator returns zero whenever a constraint is satisfied, only violated relationships contribute to the penalty.

The complete constraint-aware configuration loss is therefore defined as

\begin{equation}
    \ell_{\Theta}(\theta',\theta) =
    \ell_{\text{mse}}(\theta',\theta) +
    \lambda_{\text{pen}} \left[ \ell_{\text{range}}(\theta') + \ell_{\text{ord}}(\theta',\theta) \right],
\end{equation}

where $\theta'$ denotes either type of model output introduced above and $\lambda_{\text{pen}}\geq0$ controls the contribution of the generator-feasibility penalties. The scaler and component weights were treated as experimental hyperparameters (App.~\ref{app:ablation} describes their performance), while $\lambda_{\text{pen}}$ was set up to $1.0$ in every training run.

This discrepancy function, with the same parameterisation, was used in both objectives defined in Sec.~\ref{sec:problem:baseloss}. That is, in $\mathcal{L}_{\text{reg}}$ (Eq.~\ref{eq:loss_reg}), it compares the graph-derived prediction with the corresponding generator configuration, whereas in $\mathcal{L}_{\text{rec}}$ (Eq.~\ref{eq:loss_rec}), it compares the configuration-autoencoder reconstruction with the original configuration.

\subsection{Graph-encoder Auxiliary Clustering Objective}
\label{app:aux-cluster}

We specify below the cluster-pooling branch of the graph encoder introduced in Sec.~\ref{sec:model}, together with the auxiliary objective accompanying it. Consider a single graph $G$ and let $\mathbf{H} \in \mathbb{R}^{|V| \times m}$ collect the vertex embeddings $h_v \in \mathbb{R}^{m}$ produced by the encoder body. A learned affine map $g \colon \mathbb{R}^{m} \rightarrow \mathbb{R}^{K}$ scores every vertex against $K$ slots, and a softmax over the slot axis turns the scores into a soft assignment,

\begin{equation}
    \label{eq:aux_assign}
    \mathbf{S}_{v\kappa} = \frac{\exp\left(g(h_v)_\kappa\right)}{\sum_{\kappa'=1}^{K}\exp\left(g(h_v)_{\kappa'}\right)},
    \qquad
    \mathbf{S}\in[0,1]^{|V|\times K},
    \qquad
    \sum_{\kappa=1}^{K}\mathbf{S}_{v\kappa}=1 ,
\end{equation}

so that every vertex distributes one unit of mass over the slots. The grouping is not directly supervised: even when community information is provided as a node feature, no objective requires the learned slots to reproduce the generator-defined communities.

Each slot then receives a representation and a size,

\begin{equation}
    \label{eq:aux_readouts}
    \mathit{cl}_{e,\kappa} = \sum_{v \in V} \mathbf{S}_{v\kappa} \, h_v \in \mathbb{R}^{m},
    \qquad
    \mathit{cl}_{s,\kappa} = \sum_{v \in V} \mathbf{S}_{v\kappa},
    \qquad
    \sum_{\kappa=1}^{K}\mathit{cl}_{s,\kappa}=|V| ,
\end{equation}

where the representation sums the vertex embeddings weighted by their membership in slot $\kappa$, and the size is the mass the slot receives, which counts its vertices once the assignment is sharp. We write $\mathit{cl}_s \in \mathbb{R}^{K}$ for the vector of all $K$ slot sizes.

Neither the representations nor the sizes can be passed on as they stand, because the slot axis carries no identity: permuting the slots describes the same grouping, and two graphs may order them differently. What enters $h_G$ must therefore be invariant under such permutations. We reduce the slot representations by a component-wise maximum and the slot sizes by $Q$ of their quantiles, both of them order statistics and so invariant, and concatenate the two with the attention-pooled average,

\begin{equation}
    \label{eq:aux_hg}
    h_G = \mathrm{Proj}\left(
      \left[\,
        \textsc{AttPool}(\mathbf{H}) \; ; \;
        \max_{\kappa \leq K} \mathit{cl}_{e,\kappa} \; ; \;
        Q_{\alpha_1}\!\left(\mathit{cl}_s\right),\dots,Q_{\alpha_Q}\!\left(\mathit{cl}_s\right)
      \,\right]\right) \in \mathbb{R}^{k},
\end{equation}

where the maximum is taken component-wise, $Q_{\alpha}$ denotes the $\alpha$-quantile of the $K$ slot sizes, read at levels $\alpha_1,\dots,\alpha_Q$, and $\mathrm{Proj}$ is an affine map onto the $k$-dimensional latent space of Sec.~\ref{sec:problem}, followed by layer normalisation.

Among the three terms in Eq.~\ref{eq:total_loss_theory}, only $\mathcal{L}_{\mathrm{reg}}$ and $\mathcal{L}_{\mathrm{con}}$ depend on the soft assignments $\mathbf{S}$; the reconstruction term is confined to the configuration tower. Moreover, the assignments affect these two losses only through the component-wise maximum over the slot representations and the quantiles of the slot sizes. Neither loss directly encourages each vertex to concentrate its assignment on a single slot. Take the uniform one, $\mathbf{S}_{v\kappa}=1/K$ for every vertex and slot. All $K$ slot representations are then equal to $\frac{1}{K}\sum_{v \in V} h_v$, so the maximum returns a rescaled average over the vertices and duplicates what the attention branch already supplies. All $K$ slot sizes are equal to $|V|/K$, so the quantiles coincide and between them report nothing beyond $|V|$, which is the parameter $n$ the model has to predict in any case. The branch is then indistinguishable from a second average pool, and the distributional information it was introduced to carry would never reach $h_G$.

The auxiliary term supplies the missing preference, acting on the entropy of the assignment. For a graph whose assignment is $\mathbf{S}$, let

\begin{equation}
    \label{eq:aux_entropy}
    \eta(\mathbf{S}) =
    - \sum_{v \in V}
    \sum_{\kappa=1}^{K}
    \mathbf{S}_{v\kappa}\log \mathbf{S}_{v\kappa} ,
\end{equation}

which vanishes when every vertex commits to a single slot and is largest, $|V|\log K$, under the uniform assignment. Averaging it over every vertex of a mini-batch, whose configurations $i \in \mathcal{I}$ and replicas $r \in \mathcal{R}_i$ are indexed as in Sec.~\ref{sec:problem:baseloss}, gives

\begin{equation}
    \label{eq:aux_loss}
    \mathcal{L}_{\mathrm{aux}} =
    \frac{
      \sum_{i \in \mathcal{I}} \sum_{r \in \mathcal{R}_i}
      \eta\!\left(\mathbf{S}_i^{(r)}\right)
    }{
      \sum_{i \in \mathcal{I}} \sum_{r \in \mathcal{R}_i}
      \left|V_i^{(r)}\right|
    } ,
\end{equation}

where $\mathbf{S}_i^{(r)}$ and $V_i^{(r)}$ denote the assignment matrix and the vertex set of the replica $G_i^{(r)}$, each assignment being formed from the vertices of its own graph alone. The objective of Eq.~\ref{eq:total_loss_theory} accordingly gains a fourth term,

\begin{equation}
    \label{eq:total_loss_dcba}
    \mathcal{L}_{\mathrm{total}} =
    \lambda_{\mathrm{reg}} \mathcal{L}_{\mathrm{reg}} +
    \lambda_{\mathrm{rec}} \mathcal{L}_{\mathrm{rec}} +
    \lambda_{\mathrm{con}} \mathcal{L}_{\mathrm{con}} +
    \lambda_{\mathrm{aux}} \mathcal{L}_{\mathrm{aux}} .
\end{equation}

Minimising $\mathcal{L}_{\mathrm{aux}}$ drives each row of $\mathbf{S}$ towards a one-hot vector, so the slot sizes become counts of vertices. It does not decide how the vertices are divided, being equally minimal for the assignment that places all of them in one slot. How the vertices are divided among the slots is instead guided indirectly by $\mathcal{L}_{\mathrm{reg}}$ and $\mathcal{L}_{\mathrm{con}}$, which favour groupings useful for configuration recovery and representation alignment. Since the objective does not reward spreading the vertices over all $K$ slots either, $K$ bounds how many groups the model may take rather than prescribing their number, and it is set from the community structure of the dataset rather than tuned. Sampling graphs from the training split and counting their generator communities yields a distribution whose $95$-th percentile, taken with a safety margin of $1.2$, recommends $71$ slots, which we round up to the next power of two. The procedure is part of the released code, so it can be repeated for any dataset the framework is applied to. The reported configuration uses $K=128$ slots, $Q=15$ quantile levels spanning $0.01$ to $0.99$ and $\lambda_{\mathrm{aux}}=0.1$. Encoder variants without this branch are compared in App.~\ref{app:ablation}.

We additionally verify that the selected $K$ is not a binding constraint by measuring the effective number of occupied slots, defined as the exponential of the entropy of the slot-occupancy distribution $\mathbf{cl}_s/|V|$. Its geometric mean over the test split is $1.00$ for \texttt{DCBA} and $1.51$ for both community-aware variants, compared with $K=128$. Thus, the learned assignments use only a small fraction of the available slot capacity, suggesting that increasing $K$ would not provide additional useful capacity in the reported setting. The concentrated assignments also provide some intuition for the role of this branch. With nearly one-hot assignments, each occupied slot representation approaches an aggregate of the embeddings of its assigned vertices. Unlike attention pooling, whose normalisation tends to suppress information about absolute group size, such aggregation retains count-sensitive information, while the subsequent quantile and maximum operators capture variation and extremes across groups. This is consistent with the improvements observed in Tab.~\ref{tab:encoders}: the clearest gain is for $c_{\max}$, which directly describes an extreme of the community-size distribution, while $n_{\text{out}}$ is likewise an absolute-count parameter. The improvement in $\xi$ has a different interpretation: although $\xi$ is not count-based, it controls inter-community mixing and therefore affects how clearly the node representations can be partitioned into distinct groups.

\section{Dataset Generation Details}
\label{app:dataset-generation}

This appendix provides readers with details on how the dataset was generated and contains useful information for reproducing it or adapting the pipelines to their own needs.

\paragraph{Parameter Conversion and Validation} Parameters expressed as fractions of the graph size $n$ were converted to integer-valued generator parameters according to

\begin{align*}
    d_{\min} &= \max\big(1,\ \operatorname{round}(f_{d_{\min}} \cdot n)\big), &
    d_{\max} &= \max\big(d_{\min}+1,\ \operatorname{round}(f_{d_{\max}} \cdot n)\big), \\
    c_{\min} &= \max\big(1,\ \operatorname{round}(f_{c_{\min}} \cdot n)\big), &
    c_{\max} &= \max\big(c_{\min}+1,\ \min(n,\ \operatorname{round}(f_{c_{\max}} \cdot n))\big), \\
    n_\text{out} &= \operatorname{round}(f_{n_\text{out}} \cdot n),
\end{align*}

where $f_{(\cdot)}$ denotes the sampled fraction. The bounds on $d_{\min}$ and $c_{\min}$ guarantee minimum value of $1$, and the coupling of the upper bounds to the sampled lower bounds guarantees $d_{\max} > d_{\min}$ and $c_{\min} < c_{\max} \le n$ by construction. Every validated and accepted configuration constitutes one \emph{instance} in the sense of Sec.~\ref{sec:problem}. Each instance is then used to generate 10 independent \emph{replicas}, differing only in their random seed. As a safeguard, a candidate configuration is redrawn up to 3 times and discarded if no valid draw is found.

\paragraph{Boundary Challenge Set} The additional dataset was constructed deterministically. It enumerated all $2^9=512$ combinations obtained by assigning either the lower or upper boundary value to each of the nine sampled parameters. All resulting parameter combinations satisfied the initial configuration-validation procedure.

\paragraph{Replica Generation and Failure Handling} Replica $r \in \{1,\dots,10\}$ of each instance was generated using the seed $s_0+r$, where $s_0$ denotes the run-level seed stored in the corresponding dataset configuration. Graph generation runs were subject to predefined wall-clock limits recorded in their configuration files: 240\,s for the boundary challenge set and 60\,s for the final five chunks of the principal dataset. If the generation of any replica exceeded the applicable limit or otherwise failed, the corresponding instance was excluded from the released dataset. Of the 9\,051 valid configurations passed to the generator for the principal dataset, 8\,744 complete instances were retained. Because wall-clock failures may depend on the execution environment, the experiments reported in this study use the released datasets as fixed artefacts rather than regenerating them during model training.

\paragraph{Implementation and Reproducibility} The dataset-generation pipeline, dataset configurations, and data loaders are available at \url{https://github.com/network-science-lab/DCBA-data-set}. Generated graphs are versioned using DVC and stored remotely, with access available on request. The repository contains the complete generation pipeline, including the configuration and random seeds associated with each dataset and a common entry point for executing the generation procedure. Dataset generation was performed using Python~3.12.11 and Julia~1.12.5. The Python environment was resolved and locked using \texttt{uv}, with the complete dependency specification provided in \texttt{uv.lock}. Graphs were generated using \texttt{ABCDGraphGenerator.jl}~v0.2.0 (\url{https://github.com/bkamins/ABCDGraphGenerator.jl}, tree hash \texttt{780973a}).

\section{Algorithmic Baseline Formulation}
\label{app:baseline}

As stated in Sec.~\ref{sec:method:baseline}, the algorithmic baseline is based on the na\"{\i}ve assumption that the components of $\theta$ can be recovered independently. Thus, rather than learning a joint embedding of graphs and configurations, the baseline directly estimates each coordinate of $\theta$ from graph statistics, so that its errors reflect the extent to which the individual configuration components can be recovered without modelling their interdependence.

The community assignment is either taken directly from the ground-truth labels attached to the graph (\texttt{Base-T}) or, when the ground truth is assumed to be unavailable, obtained by running the Leiden algorithm~\cite{traag2019leiden} on the graph (\texttt{Base-L}). This second mode simulates the realistic setting in which no community labels accompany the observed network and allows us to assess how sensitive the baseline is to replacing the ground-truth labels with a detected partition.

Given a graph and a community assignment, $\widehat{n}$ is the number of vertices in the graph, $\widehat{d}_{\min}$ and $\widehat{d}_{\max}$ are set to the minimum and maximum observed vertex degrees, and $\widehat{c}_{\min}$ and $\widehat{c}_{\max}$ are set to the minimum and maximum observed community sizes. The power-law exponents $\widehat{t}_1$ and $\widehat{t}_2$ are then obtained by numerically maximising the truncated power-law log-likelihood of the degree sequence and the community-size sequence, respectively. The noise level $\xi$ is estimated as the fraction of edges crossing community boundaries,

\begin{equation}
    \widehat{\xi} = 1 - \frac{e_{\mathrm{intra}}}{|E|},
\end{equation}

where $|E|$ is the number of edges in the graph and $e_{\mathrm{intra}}$ is the number of edges whose endpoints belong to the same community. The number of outliers $\widehat{n}_{\mathrm{out}}$ is estimated by scoring each vertex $v$ as

\begin{equation}
    B(v) = 2 \left( \frac{\deg_A(v)}{\deg_G(v)} - \frac{\operatorname{vol}(A) - \deg_G(v)}{\operatorname{vol}(V)} \right),
\end{equation}

where $A$ is the community containing $v$, $\deg_G(v)$ denotes the degree of $v$ in $G$, $\deg_A(v)$ denotes the number of its intra-community edges within $A$, and $\operatorname{vol}(\cdot)$ denotes the sum of degrees over a vertex set~\cite{kaminski2023abcdo}. Isolated vertices, for which $\deg(v)=0$, are excluded rather than counted as outliers.

The resulting $\widehat{\theta}$ can be directly compared with the predictions of \texttt{DCBA}. Owing to the stochastic nature of \texttt{ABCD} and the collisions and corrections occurring during graph generation, only $n$ can be recovered exactly from every realisation. The accuracy with which the baseline recovers the remaining parameters instead depends on the particular configuration being estimated.

\section{Model Selection, Ablations, and Interpretability}
\label{app:ablation}

This appendix reports the experiments used to select the model configuration evaluated in Sec.~\ref{sec:experiments} and to assess its sensitivity to key design choices. These experiments examine the graph encoder (including the way the clustering signal is processed), $\theta$-scalers, the use of ground-truth community information as node features, and the mini-batch size. They were conducted sequentially, with each subsequent experiment adopting the best-performing configuration identified in the preceding one; unless stated otherwise, only the component under investigation was varied. We additionally analyse the stability of the learned graph representations, and their sensitivity to varying parameters of \texttt{ABCD}. Finally, the implementation and reproducibility details are presented.

\subsection{Graph Encoders}
\label{app:encoder-ablation}

The first experiment compares candidate graph-encoder architectures and selects the variant used in the subsequent experiments. All variants employed the logarithmic scaler described in App.~\ref{app:loss:scaler}, a mini-batch size of $32$, and constant node features, $\mathbf{X}=\mathbf{1}$. The evaluated encoders are described below, and their performance is compared in Tab.~\ref{tab:encoders}.

\paragraph{The \texttt{GIN} Encoder} A stack of \texttt{GIN} layers~\cite{xu2018powerful}, whose outputs are combined by a trainable attention-weighted sum and aggregated into $h_G$ by attention pooling. It relies on local message passing alone and therefore serves as a reference point for what can be recovered without a graph-wide view.

\paragraph{The \texttt{GPS} Encoder} Here the \texttt{GIN} layers are replaced with the \texttt{GraphGPS} blocks of Sec.~\ref{sec:model}, in which local message passing is combined with global multi-head self-attention, implemented using the linear approximation of~\cite{choromanski2021rethinking}. The readout remains a single attention pooling, so the graph is summarised by a weighted average of its vertex embeddings.

\paragraph{The \texttt{GPS-weighted} Encoder} The architecture is left unchanged and only the configuration loss is modified, each parameter receiving a weight $w_i = 1 + \operatorname{clip}(R^2_{\text{regr}} - R^2_{\text{cross}}, 0, 1)$. The weight measures how much of $\theta_i$ is lost when it is decoded from $z_G$ rather than from $z_{\theta}$, so that the gradient is directed towards the parameters recovered worst by the graph encoder rather than by the configuration autoencoder. Under this rule $t_2$ receives the largest weight, close to $2$.

\paragraph{The \texttt{GPS-clustered} Encoder} This variant adds the soft-clustering branch of Sec.~\ref{sec:model} alongside the attention pooling. Vertices are softly assigned to at most $K = 128$ learned slots, and quantiles of the resulting slot sizes are appended to $h_G$. The slots receive no supervision, as ground-truth communities constitute generator metadata that would be unavailable for an observed graph.

\paragraph{The \texttt{GPS-clustered-weighted} Encoder} Both extensions are applied together, that is the cluster-pooling branch trained under the weighted configuration loss. This is the encoder presented in Sec.~\ref{sec:model} and used by every \texttt{DCBA} variant evaluated in Sec.~\ref{sec:experiments}.

\begin{table}[ht!]
    \centering
    \caption{Parameter-wise $R^2$ on the casual test set for the compared graph encoders; first- and second-best results are shown in green.}
    \label{tab:encoders}
    \small
    \begin{tabular}{@{}lrrrrr@{}}
        \toprule
        Parameter & \texttt{GIN} & \texttt{GPS} & \texttt{GPS-weighted} & \texttt{GPS-clustered} & \texttt{GPS-clustered-weighted} \\
        \midrule
        $n$ & 0.36 & \second{0.93} & 0.86 & \first{0.95} & 0.89 \\
        $t_1$ & 0.68 & \first{0.95} & \second{0.94} & 0.87 & \first{0.95} \\
        $d_{\min}$ & 0.84 & 0.92 & \second{0.94} & \first{0.96} & \first{0.96} \\
        $d_{\max}$ & 0.01 & 0.74 & \second{0.76} & \first{0.78} & \first{0.78} \\
        $t_2$ & \second{-0.02} & \first{0.00} & -0.05 & \second{-0.02} & \second{-0.02} \\
        $c_{\min}$ & 0.13 & \first{0.46} & 0.15 & \second{0.44} & \second{0.44} \\
        $c_{\max}$ & 0.17 & 0.37 & 0.21 & \first{0.52} & \second{0.45} \\
        $\xi$ & 0.10 & 0.79 & \second{0.82} & \first{0.85} & \first{0.85} \\
        $n_{\text{out}}$ & 0.10 & 0.15 & 0.13 & \first{0.27} & \second{0.24} \\
        \midrule
        $\overline{R^2}$ & 0.26 (0.31) & 0.59 (0.36) & 0.53 (0.41) & \first{0.62} (0.34) & \second{0.61} (0.35) \\
        \bottomrule
    \end{tabular}
\end{table}

The \texttt{GIN} encoder performs considerably worse than all \texttt{GPS}-based variants, indicating the importance of incorporating global interactions in addition to local message passing. Adding the weighted pooling branch to \texttt{GPS} has a mixed effect, improving several parameters but reducing the overall $\overline{R^2}$. In contrast, the soft-clustering branch provides the most consistent improvement, particularly for $c_{\max}$, $\xi$, and $n_{\mathrm{out}}$, and yields the highest average score. Combining both aggregation mechanisms largely preserves these gains while recovering the performance on $t_1$. As only $0.01$ in $\overline{R^2}$ separates \texttt{GPS-clustered} and \texttt{GPS-clustered-weighted}, we use the latter in the remaining experiments.

\subsection{Representing Cluster Structure}

This experiment compares different hyperparameters of the soft-clustering module presented in Sec.~\ref{sec:model} and App.~\ref{app:aux-cluster}. All variants follow the base configuration from App.~\ref{app:encoder-ablation} and use \texttt{GPS-clustered-weighted}, varying only the \texttt{Q} and \texttt{K} hyperparameters. The selected variants for this comparison are described below, and their results are shown in Tab.~\ref{tab:ablation-k} and Tab.~\ref{tab:ablation-q}.

\paragraph{The \texttt{K} Ablation} The number of learned slots in the soft-clustering pool is varied across $K \in \{64, 128, 256\}$. The model can leave some slots unused (as $K$ is a capacity budget only), and the \texttt{Q-wide} readout is used in this case, i.e., the same as reported in App.~\ref{app:aux-cluster}.

\begin{table}[ht!]  
    \centering          
    \caption{Parameter-wise $R^2$ on the casual test set for varying numbers of clusters $K$; first- and second-best results are shown in green.}     
    \label{tab:ablation-k}
    \small
    \begin{tabular}{@{}lrrr@{}}
        \toprule
        Parameter & \texttt{K-64} & \texttt{K-128} & \texttt{K-256} \\
        \midrule
        $n$ & \second{0.86} & \first{0.89} & 0.81 \\
        $t_1$ & \second{0.94} & \first{0.95} & 0.92 \\
        $d_{\min}$ & \second{0.95} & \first{0.96} & 0.94 \\
        $d_{\max}$ & 0.70 & \first{0.78} & \second{0.75} \\
        $t_2$ & \first{0.01} & \second{-0.02} & \first{0.01} \\
        $c_{\min}$ & 0.32 & \first{0.44} & \second{0.38} \\
        $c_{\max}$ & \second{0.33} & \first{0.45} & 0.29 \\
        $\xi$ & \second{0.82} & \first{0.85} & \first{0.85} \\
        $n_{\text{out}}$ & 0.06 & \first{0.24} & \second{0.15} \\
        \midrule
        $\overline{R^2}$ & 0.55 (0.36) & \first{0.61} (0.35) & \second{0.57} (0.34) \\
        \bottomrule
    \end{tabular}
\end{table}

\paragraph{The \texttt{Q-wide} Readout} A dense set of $15$ hand-picked quantile levels is used, covering both size extremes and the bulk of the distribution: $\{ 0.01, 0.05, 0.1, 0.2, 0.25, 0.3, 0.4, 0.5, 0.6, 0.7, 0.75, 0.8, 0.9, 0.95, 0.99 \}$. This provides a detailed representation of the community-size distribution across the full range. The soft-clustering pool uses $K = 128$ learned slots.

\paragraph{The \texttt{Q-0} Readout} No size-distribution information is used. The encoder keeps only the max-pooled cluster embedding, without an explicit representation of how community sizes are distributed.

\paragraph{The \texttt{Q-coarse} Readout} A coarse four-point quantile summary is used, with quantile levels $\{0.25, 0.5, 0.75, 0.99\}$. This provides a compact representation combining information from the center and upper tail of the distribution.

\paragraph{The \texttt{Q-top} Readout} Only the upper tail of the community-size distribution is represented, using quantile levels $\{0.9, 0.99, 0.999, 1\}$. This focuses on the largest communities while discarding information from the bulk and lower end of the distribution.

\paragraph{The \texttt{Q-uniform} Readout} A set of 15 quantile levels is spaced uniformly across the full range, using $\mathrm{linspace}(0,1,15)$. This provides an even representation of the community-size distribution without concentrating levels in particular regions.

\begin{table}[ht!]
    \centering
    \caption{Parameter-wise $R^2$ on the casual test set for varying cluster-size quantile sets $Q$; first- and second-best results are shown in green.}
    \label{tab:ablation-q}
    \small
    \begin{tabular}{@{}lrrrrr@{}}
        \toprule
        Parameter & \texttt{Q-wide} & \texttt{Q-0} & \texttt{Q-coarse} & \texttt{Q-top} & \texttt{Q-uniform} \\
        \midrule
        $n$ & 0.89 & \second{0.92} & \first{0.94} & \second{0.92} & \first{0.94} \\
        $t_1$ & \first{0.95} & \second{0.94} & \first{0.95} & \first{0.95} & 0.92 \\
        $d_{\min}$ & \first{0.96} & 0.94 & \second{0.95} & \second{0.95} & 0.93 \\
        $d_{\max}$ & 0.78 & \second{0.79} & \first{0.82} & 0.78 & 0.73 \\
        $t_2$ & \second{-0.02} & \second{-0.02} & \second{-0.02} & \first{-0.01} & \first{-0.01} \\
        $c_{\min}$ & 0.44 & \second{0.45} & \first{0.50} & 0.37 & 0.39 \\
        $c_{\max}$ & \second{0.45} & 0.38 & \first{0.56} & 0.37 & 0.41 \\
        $\xi$ & \first{0.85} & \first{0.85} & \first{0.85} & \second{0.81} & 0.75 \\
        $n_{\text{out}}$ & \second{0.24} & \first{0.33} & \first{0.33} & 0.23 & 0.22 \\
        \midrule
        $\overline{R^2}$ & 0.61 (0.35) & \second{0.62} (0.33) & \first{0.65} (0.32) & 0.60 (0.34) & 0.59 (0.33) \\
        \bottomrule
    \end{tabular}
\end{table}

Table~\ref{tab:ablation-k} shows a non-monotonic trend in the number of clusters $K$, The default budget of $128$ slots achieves the highest mean $R^2$ ($0.61$), while both $K=64$ ($0.55$) and $K=256$ ($0.57$) underperform it consistently. This supports $K=128$ as a well-calibrated choice. The picture for the quantile set $Q$ presented in Tab.~\ref{tab:ablation-q} is less straightforward. The four-point \texttt{Q-coarse} configuration achieves the highest mean $R^2$ ($0.65$), followed by \texttt{Q-0} ($0.62$) and \texttt{Q-wide} ($0.61$), while \texttt{Q-top} ($0.60$) and \texttt{Q-uniform} ($0.59$) perform worse. This suggests that most of the exploitable size-distribution signal is captured by a small number of well-placed quantiles. We nonetheless retain \texttt{Q-wide} as the default configuration for consistency with the main experiments reported in the paper. The additional ablation shows that \texttt{Q-coarse} provides a strong alternative, and may be preferable in follow-up studies, while leaving the conclusions regarding the utility of preserving the clustering signal unchanged.

\subsection{Configuration Scalers}

The second experiment evaluates alternative methods for scaling the generator configurations. As the components of $\theta$ differ substantially in magnitude and interpretation, the choice of scaler may affect both regression and contrastive learning. We were mainly interested in maps built from the generator's own constraints rather than from statistics fitted to the training data, which would tie a trained model to the ranges the dataset was sampled from. The compared scalers are described below. The results are reported in Tab.~\ref{tab:scalers}.

\begin{table}[ht!]
    \centering
    \caption{Parameter-wise $R^2$ on the casual test set for the compared configuration scalers; first- and second-best results are shown in green.}
    \label{tab:scalers}
    \small
    \begin{tabular}{@{}lrrrrr@{}}
        \toprule
        Parameter & \texttt{identity} & \texttt{n-max} & \texttt{relative} & \texttt{log} & \texttt{empirical} \\
        \midrule
        $n$ & \first{0.98} & \first{0.98} & \first{0.98} & \second{0.89} & \first{0.98} \\
        $t_1$ & -11.72 & 0.78 & 0.87 & \first{0.95} & \second{0.90} \\
        $d_{\min}$ & \second{0.32} & -1.26 & -0.32 & \first{0.96} & -0.20 \\
        $d_{\max}$ & 0.80 & \second{0.81} & \first{0.82} & 0.78 & 0.80 \\
        $t_2$ & -6.97 & \second{-0.04} & \second{-0.04} & \first{-0.02} & \first{-0.02} \\
        $c_{\min}$ & \second{0.31} & 0.29 & -0.03 & \first{0.44} & 0.13 \\
        $c_{\max}$ & 0.45 & 0.42 & \second{0.51} & 0.45 & \first{0.55} \\
        $\xi$ & -0.37 & 0.73 & 0.75 & \first{0.85} & \second{0.83} \\
        $n_{\text{out}}$ & \first{0.39} & 0.36 & \second{0.38} & 0.24 & 0.33 \\
        \midrule
        $\overline{R^2}$ & -1.76 (4.48) & 0.34 (0.68) & 0.44 (0.47) & \first{0.61} (0.35) & \second{0.48} (0.44) \\
        \bottomrule
    \end{tabular}
\end{table}

\paragraph{The \texttt{identity} Scaler} No transformation is applied, so that $\theta$ is presented to the model in its raw units. As $n$ reaches several thousand while $t_1$, $t_2$ and $\xi$ remain confined to narrow ranges, it illustrates the effect of leaving the differences in magnitude unaddressed.

\paragraph{The \texttt{n-max} Scaler} A single linear map per feature, with bounds anchored on the largest admissible $n$ rather than on the sampling ranges of Tab.~\ref{tab:abcdo-params}. It is the simplest of the compared transformations, at the cost of compressing the small-valued parameters into a narrow band near zero.

\paragraph{The \texttt{relative} Scaler} Every constrained feature becomes a fraction of the parameter bounding it, namely $c_{\max}/n$, $d_{\max}/n$, $n_{\text{out}}/n$, $c_{\min}/c_{\max}$, and $d_{\min}/d_{\max}$, while $n$, $t_1$, $t_2$, and $\xi$ remain scaled linearly by $10000$, $5$, and $5$, respectively. The ordering constraints are then satisfied by construction whenever the outputs lie within $[0, 1]$.

\paragraph{The \texttt{empirical} Scaler} Min-max scaling from the sampling ranges reported in Tab.~\ref{tab:abcdo-params}, with the exception of $c_{\min}$ and $d_{\min}$, which are again expressed relative to their constraint partners.

The \texttt{log} scaler attains the best result for most parameters and by a wide margin in terms of $\overline{R^2}$, and is therefore retained in the remaining experiments. The alternatives recover $n$ almost exactly, which is expected as all of them keep it on a linear scale, but they lose $d_{\min}$ entirely. The \texttt{identity} scaler illustrates the consequence of leaving the raw magnitudes untouched, as the regression becomes dominated by $n$ while both power-law exponents deteriorate.

\subsection{Node Features}

As described in Sec.~\ref{sec:model}, the graph encoder includes a soft-clustering branch intended to capture group-level structure. We additionally evaluated whether providing ground-truth community information as node features improves model performance. To preserve invariance to permutations of community labels, each vertex was assigned the size of its ground-truth community rather than its community identifier. Table~\ref{tab:features} compares this setting with the baseline using constant node features, $\mathbf{X}=\mathbf{1}$.

\begin{table}[ht!]
    \centering
    \caption{Parameter-wise $R^2$ on the casual test set for constant ($\mathbf{X}=\mathbf{1}$) and community-size node features; first- and second-best results are shown in green.}
    \label{tab:features}
    \small
    \begin{tabular}{@{}lrr@{}}
        \toprule
        Parameter & \texttt{base} & \texttt{comm} \\
        \midrule
        $n$ & \second{0.89} & \first{0.91} \\
        $t_1$ & \first{0.95} & \second{0.92} \\
        $d_{\min}$ & \first{0.96} & \first{0.96} \\
        $d_{\max}$ & \first{0.78} & \first{0.78} \\
        $t_2$ & \second{-0.02} & \first{0.07} \\
        $c_{\min}$ & \second{0.44} & \first{0.80} \\
        $c_{\max}$ & \second{0.45} & \first{0.71} \\
        $\xi$ & \second{0.85} & \first{0.97} \\
        $n_{\text{out}}$ & \second{0.24} & \first{0.34} \\
        \midrule
        $\overline{R^2}$ & \second{0.61} (0.35) & \first{0.72} (0.31) \\
        \bottomrule
    \end{tabular}
\end{table}

Community sizes improve precisely the parameters they describe, raising $c_{\min}$ and $c_{\max}$ by approximately $0.3$ each and $\xi$ to $0.97$, with only a slight loss on $t_1$. The information comes from the ground-truth assignment rather than from a community detection algorithm, which is the reason for pairing this variant with \texttt{Base-T} in Sec.~\ref{sec:experiments}.

\subsection{Batch Size}

The next experiment examines whether the mini-batch size affects model performance. Following common practice in contrastive learning, we expected larger mini-batches to improve representation learning by exposing the model to more positive and negative pairs within each update. The results are reported in Tab.~\ref{tab:batch}.

\begin{table}[ht!]
    \centering
    \caption{Parameter-wise $R^2$ on the casual test set for the compared mini-batch sizes; first- and second-best results are shown in green.}
    \label{tab:batch}
    \small
    \begin{tabular}{@{}lrr@{}}
        \toprule
        Parameter & \texttt{Batch 32} & \texttt{Batch 64} \\
        \midrule
        $n$ & \second{0.91} & \first{0.94} \\
        $t_1$ & \second{0.92} & \first{0.94} \\
        $d_{\min}$ & \first{0.96} & \first{0.96} \\
        $d_{\max}$ & \first{0.78} & \second{0.74} \\
        $t_2$ & \second{0.07} & \first{0.13} \\
        $c_{\min}$ & \second{0.80} & \first{0.86} \\
        $c_{\max}$ & \second{0.71} & \first{0.75} \\
        $\xi$ & \first{0.97} & \first{0.97} \\
        $n_{\text{out}}$ & \first{0.34} & \second{0.33} \\
        \midrule
        $\overline{R^2}$ & \second{0.72} (0.31) & \first{0.74} (0.31) \\
        \bottomrule
    \end{tabular}
\end{table}

Doubling the mini-batch improves most parameters after 20 epochs, which was the maximum training duration used in the ablation study, with per-parameter differences ranging from $-0.03$ to $+0.06$ and the largest gains observed for $c_{\min}$, $t_2$, and $c_{\max}$. However, after extending training to 50 epochs (Tab.~\ref{tab:r2-results}), this advantage largely disappears. This suggests that the larger batch primarily accelerates convergence by exposing the contrastive objective to more positive and negative pairs per update, rather than yielding a consistently better final solution.

\subsection{Stability of Graph Representations}

Finally, we present an additional analysis of the embedding stability for the \texttt{Batch 64} variant (described in Tab.~\ref{tab:batch}). From the trained model, we retrieve $z_G$ embeddings of test graphs along with their ABCD parameters $\theta$. Then, an Euclidean distance between a subset of pairs of embeddings is calculated:

\begin{equation}
 d(z_G)=d(z_{G,i},z_{G,j})=\lVert z_i - z_j \rVert_2.
\end{equation}

\begin{figure}[ht]
    \centering
    \includegraphics[width=0.55\textwidth]{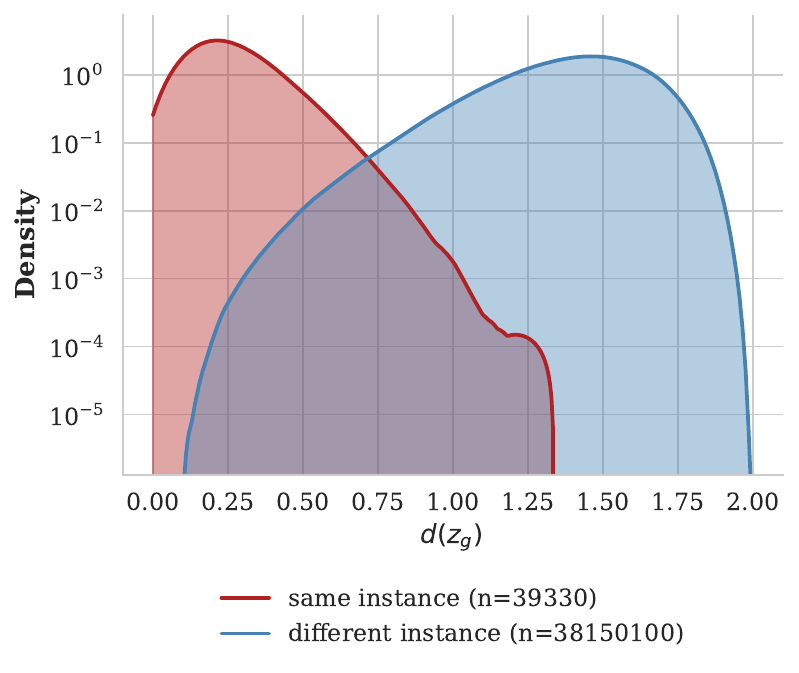}
    \caption{Distribution of pairwise $z_G$ embedding distances for same-instance vs.\ cross-instance replica pairs, log-scaled density.}
    \label{fig:embed-stability}
\end{figure}

Pairs are split into \emph{same-instance} replicas ($\theta_i=\theta_j$) and \emph{cross-instance} ($\theta_i \ne \theta_j$) sets. Figure~\ref{fig:embed-stability} shows that the same-instance distance distribution is concentrated at small distances, consistent with embedding stability across replicas of the same underlying instance. The cross-instance distribution is shifted towards larger distances and is considerably more diffuse, reflecting greater embedding dissimilarity between distinct instances.

\begin{figure}[ht]
    \centering
    \includegraphics[width=0.45\textwidth]{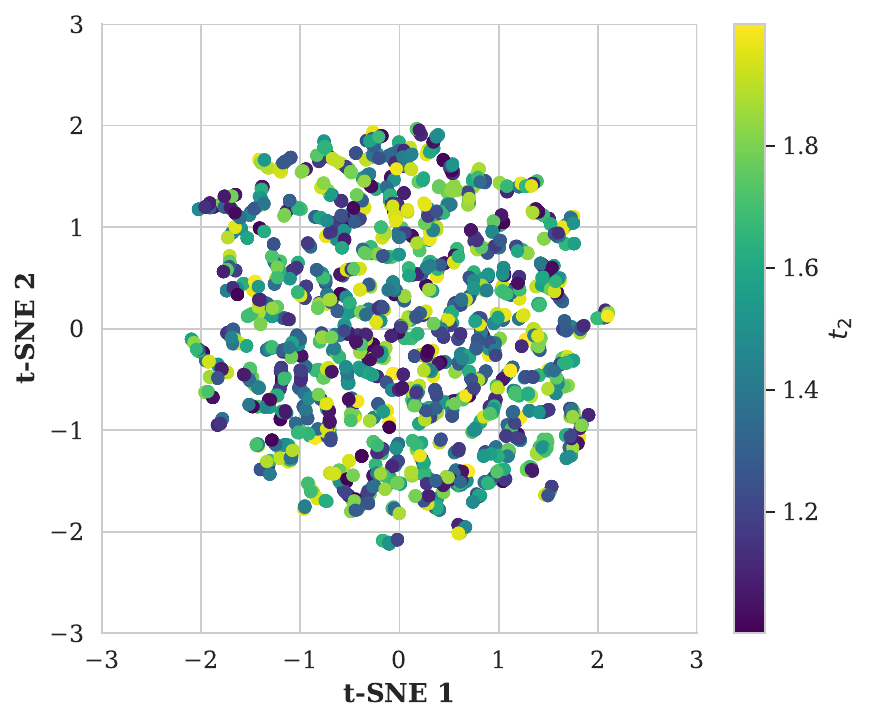}
    \includegraphics[width=0.45\textwidth]{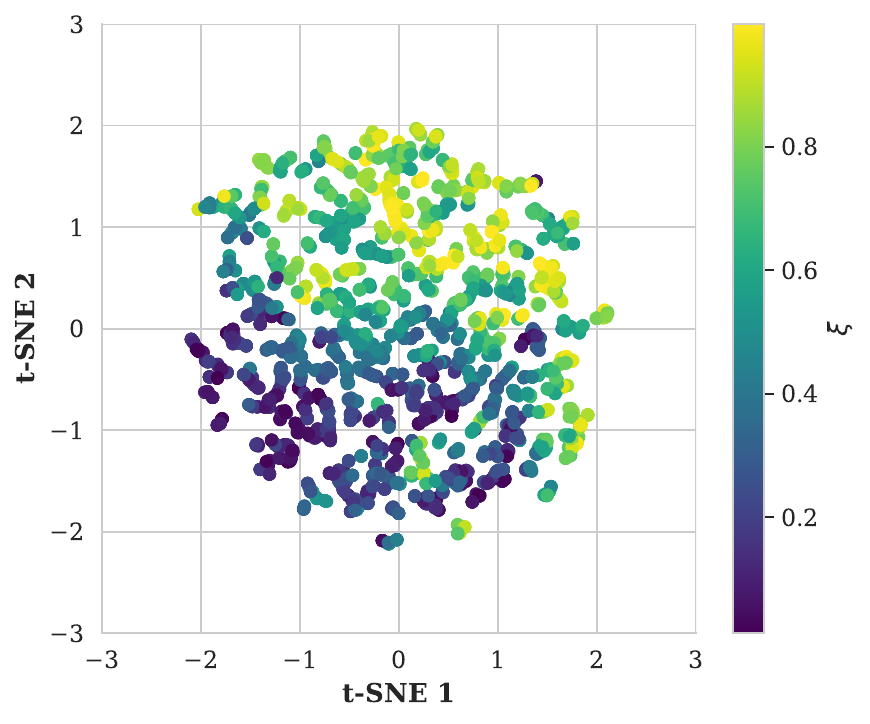}
    \caption{$\mathbb{R}^2$ representation of $z_G$ with $t_2$ (left) and $\xi$ (right) colour overlay.}
    \label{fig:embed-mosaic}
\end{figure}

To check the stability of embeddings' geometry with respect to each $\theta$ parameter, we project $z_G$ embeddings from $\mathbb{R}^{64}$ to $\mathbb{R}^2$ space with the t-SNE algorithm. Colouring by the raw values of an ABCD parameter provides guidance as to whether the learned space visibly organises along a parameter. Figure~\ref{fig:embed-mosaic} presents a good and bad fit for two selected parameters. Values of the community sizes exponent $t_2$ are scrambled within the low-dimensional space, suggesting weak performance of the model in $t_2$ retrieval (see Tab.~\ref{tab:r2-results}). In contrast, high values of $\xi$ are well concentrated in one area, pointing to higher performance for this parameter.

\subsection{Latent-space Sensitivity to Generator Parameters}

To gain further understanding of the relationship between \texttt{ABCD}'s parameters and the learned space of graph embeddings obtained for the \texttt{Batch 32} variant from Tab.~\ref{tab:batch}, we performed a distance metric learning task. The goal of the experiment is to better understand the inner workings of \texttt{DCBA} in the spirit of eXplainable AI. Although the results are not directly related to model quality, they can shed light on the importance of the synthetic graph generator's parameters. Let $M_G$ and $M_{\theta}$ be the Mahalanobis~\cite{mahalanobis1936generalized} matrices defining the distances:

\begin{equation}
    d_G(z_{G,i}, z_{G,j}) = \sqrt{(z_{G,i} - z_{G,j})^\top M_G (z_{G,i} - z_{G,j})}, \quad d_{\theta}(\theta_i, \theta_j) = \sqrt{(\theta_i - \theta_j)^\top M_{\theta} (\theta_i - \theta_j)}
\end{equation}

where $z_{G,i}$ is the $i$-th normalised graph embedding and $\theta_i$ is the min--max scaled vector of the $i$-th graph's \texttt{ABCD} parameters. We fitted $M_G$ and $M_{\theta}$ to minimise the pairwise difference between distances in the two spaces using a scale-invariant MSE loss. Intuitively, we seek linear transformations, defined by $M_G$ and $M_{\theta}$, such that if two points are close in the embedding space, they are also close in the \texttt{ABCD} parameter space. Then, via eigendecomposition of $M_{\theta}$, we can infer the sensitivity of the network's learned geometry to individual \texttt{ABCD} parameters.

\begin{figure}[ht]
    \centering
    \includegraphics[width=0.9\textwidth]{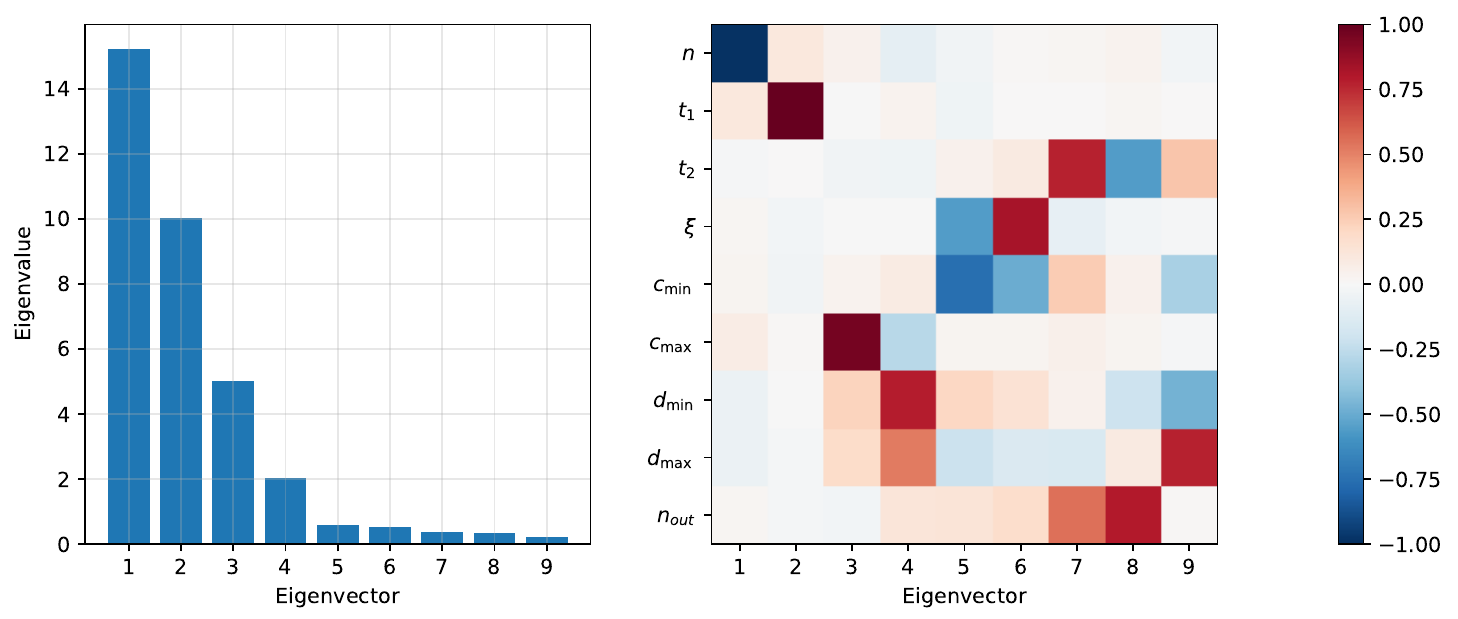}
    \caption{Eigenvalue spectrum (left) and per-parameter eigenvector loadings (right) of $M_{\theta}$ matrix.}
    \label{fig:b-summary}
\end{figure}

With the optimal Mahalanobis matrices, distances in the two spaces on held-out data correlate strongly (Pearson $r=0.867$), indicating that embedding-space proximity reflects closeness in the underlying generative parameters under a linear-transformation regime. Figure~\ref{fig:b-summary} shows the eigenvalue spectrum concentrated in three components -- $n$, $t_1$, and $c_{\max}$ -- indicating their importance in shaping the embedding manifold. This suggests the model can capture diverse topological features, i.e., graph size ($n$), degree structure ($t_1$), and a community-related characteristic ($c_{\max}$). Graph size, a basic characteristic, is expected to have a profound impact on the discriminative capabilities of the inverse model, but the vital role of $t_1$ and $c_{\max}$ is not as easily anticipated.

\subsection{Implementation and Reproducibility}

The source code of the training pipeline is available at \url{https://github.com/network-science-lab/DCBA-trainer}. The implementation was developed in Python 3.12 using PyTorch 2.5.1. All \texttt{DCBA} variants were optimised using AdamW (learning rate $0.001$, weight decay $0.00001$) and trained with a fixed random seed. The experiments presented in the main part of the manuscript were run for at most 50 epochs with an early-stopping patience of 10, whereas the ablations reported in the appendices were run for at most 20 epochs with a patience of 3. The former required on average 27.4 epochs and the latter 12.6 epochs, with one epoch on the ``casual'' dataset taking on average 4.1 hours. The experiments were conducted on two machines, both equipped with an Intel Xeon Gold 6330 CPU and 64 GB of RAM. One featured an NVIDIA TITAN RTX GPU, while the other, used for the experiments with batch size 64, with an NVIDIA A40 GPU.

\section{Author Contributions}

\textbf{Piotr Br{\'o}dka}: Methodology; Resources; Writing -- Review \& Editing; Supervision; Funding acquisition. \textbf{Micha{\l} Czuba}: Conceptualisation; Methodology; Software; Validation; Formal analysis; Investigation; Data Curation; Writing -- Original Draft; Writing -- Review \& Editing; Visualisation; Project Administration. \textbf{Bogumi\l{} Kami\'{n}ski}: Methodology; Supervision; Funding acquisition. \textbf{\L{}ukasz Krai\'{n}ski}: Methodology; Software; Validation; Formal analysis; Investigation; Data Curation; Writing -- Original Draft; Writing -- Review \& Editing; Visualisation. \textbf{Katarzyna Musial}: Supervision. \textbf{Pawe\l{} Pra\l{}at}: Supervision. \textbf{Mateusz Stolarski}: Methodology; Software; Validation; Formal analysis; Investigation; Data Curation; Writing -- Original Draft; Writing -- Review \& Editing; Visualisation.\footnote{The statement has been prepared in accordance with the CRediT taxonomy (\url{https://www.elsevier.com/researcher/author/policies-and-guidelines/credit-author-statement}).}


\paragraph{AI Disclosure} In this work, generative AI tools were used for spelling corrections in the manuscript and to facilitate source-code development at the level of implementing the authors' ideas. The authors take full responsibility for the final content of this work. No creative work was commissioned to AI.

\end{document}